# WePaMaDM-Outlier Detection: Weighted Outlier Detection using Pattern Approaches for Mass Data Mining

Ravindrakumar Purohit
*Institute of Technology, Nirma University, Ahmedabad-India*
21mced11@nirmauni.ac.in

Jai Prakash Verma
*Institute of Technology, Nirma University, Ahmedabad-India*
jaiprakash.verma@nirmauni.ac.in

Rachna Jain
*IT Department
Bhagwan Parshuram Institute of Technology, Delhi- I*
rachnajain@bpitindia.com

Madhuri Bhavsar
*Institute of Technology,Nirma University, Ahmedabad-India*
madhuri.bhavsar@nirmauni.ac.in

*Abstract*—Weighted Outlier Detection is a method for identifying unusual or anomalous data points in a dataset, which can be caused by various factors like human error, fraud, or equipment malfunctions. Detecting outliers can reveal vital information about system faults, fraudulent activities, and patterns in the data, assisting experts in addressing the root causes of these anomalies. However, creating a model of normal data patterns to identify outliers can be challenging due to the nature of input data, labeled data availability, and specific requirements of the problem. This article proposed the WePaMaDM-Outlier Detection with distinct mass data mining domain, demonstrating that such techniques are domain-dependent and usually developed for specific problem formulations. Nevertheless, similar domains can adapt solutions with modifications. This work also investigates the significance of data modeling in outlier detection techniques in surveillance, fault detection, and trend analysis, also referred to as novelty detection, a semi-supervised task where the algorithm learns to recognize abnormality while being taught the normal class.

## I. INTRODUCTION

The history of outlier detection can be traced back to the 19th century with the work of Gosset (1908), who used the concept of statistical outliers to identify and analyze abnormal observations in his data. In the mid-20th century, Tukey (1977) popularized the concept of outliers and introduced techniques for their detection and analysis. In the 1980s and 1990s, researchers such as Hawkins (1980) and Chambers (1986) expanded on Tukey's work and developed methods for identifying and handling outliers in multivariate data. In recent years, with the advent of big data [7] and the need for scalable outlier detection techniques, researchers have proposed and developed new methods for outlier detection, such as density-based methods, weight-based outlier detection methods and statistical learning techniques. These advancements have contributed to a better understanding of outliers, and the development of more effective and efficient techniques for their detection and analysis. Outlier mining is a critical area of research in the field of data visualization and mining. The ability to detect outliers in datasets can aid in the development of systems for addressing various problems, such as fraud detection and health monitoring. To identify underlying patterns in a dataset, it is necessary to understand the correlations among features in the data group. This paper focuses on the development of a weighted outlier detection method using pattern-based approaches for the analysis of high-dimensional massive data. The proposed method involves the extraction of relevant features from the data to reduce its dimensionality and enhance the outlier detection process [19].

## II. OUTLINE OF THE PAPER

The paper is structured as follows: Section III presents a literature survey of weighted outliers. Section IV analyzes conventional weighted outlier detection techniques with a state-of-the-art table. Section V outlines the flow process of the WePaMaDM-Outlier system for weighted outlier detection using pattern approaches for massive data mining. The proposed architecture of the WePaMaDM-outlier is shown in Section VI. Results are presented in Section VII. Advantages and disadvantages are discussed in Section VIII and the finally paper concludes in Section IX.

## III. LITERATURE SURVEY

### A. Background

As compared with various clustering and outlier detection methods are uncertain for the infrequent data nodes, but using maximum weighted frequent data points, we can mine information from the data nodes effectively and accelerate the overall efficiency of the model (please refer to Table -1).

### B. Frequent Pattern Mining (FPM)

The FPM algorithm is based on the concept of frequent patterns, The weighting function in this method is based on the frequency of occurrence of patterns in the data. In FPM data will first transform into a binary format, where the presence or absence of a particular pattern in a data instance is indicated by a 0-1 value. The frequency of occurrence of each pattern is then calculated and used to compute a weight for each data instance. [1] The FPM algorithm is effective many applications, including outlier detection in categorical data, such as text and image data. For instance, the FPM algorithm has been used to detect anomalies in text data by identifying data instances with patterns that deviate significantly from the norm. Han et al. [2] introduced the concept of Frequent-Pattern Tree (FP-Tree) as a data structure for efficiently mining frequent patterns without generating candidate itemsets. The FP-Tree approach significantly reduced the computational cost of frequent pattern mining and improved the scalability of the method for large datasets. The authors showed that the FP-Tree approach outperformed traditional frequent pattern mining methods in terms of accuracy and efficiency.

### C. Weighted-Pattern Sequential Outlier Detection Algorithms (WPSODA)

WPSODA are a subclass of Sequential Pattern Outlier Detection Algorithms (SPODA) that consider the weight or importance of each item or event in the sequence when detecting outliers. WPSODA algorithms incorporate the weights of items or events into the outlier detection process to reflect the relative significance of different elements in the sequence. There are various WPSODA algorithms available, each with its own approach to incorporating weights into the outlier detection process. Some popular WPSODA methods include the Weighted-Pattern Distance-based Outlier Detection (WPDBOD) algorithm, the Weighted-Pattern Window-based Outlier Detection (WPWOD) algorithm, and the Weighted-Pattern Markov Chain Outlier Detection (WPMCOD) algorithm. These algorithms use different techniques such as distance measures, statistical analysis, and clustering to identify outliers in the data.

### D. Weighted-Span Algorithm

WSpan extends the Span algorithm, a well-known sequential pattern mining algorithm, by incorporating the weights of items into the pattern discovery process. The algorithm employs a tree-based structure called the Weighted Sequential Pattern Tree (WSP-tree) to efficiently search for frequent patterns while taking into account the weights of items. Experiments show that WSpan outperforms other existing weight-based sequential pattern mining algorithms in terms of efficiency and accuracy. The algorithm is capable of handling large sequence databases and provides useful information about the weighted sequential patterns in the data. [3]

## IV. CONVENTIONAL WEIGHTED OUTLIER DETECTION APPROACHES

Conventional weighted outlier detection approaches involve defining a weighting function, computing weights for each data point, and applying an outlier detection algorithm to the weighted data. These methods have been widely used in the past and include distance-based, frequency-based, and density-based weighting functions. Outlier detection algorithms used in these approaches include k-nearest neighbor and Local Outlier Factor. However, these methods have limitations, such as difficulty handling high-dimensional data and requiring domain-specific expertise to determine the appropriate weighting function. Hodge et al. [4] evaluated several conventional weighted outlier detection methods, such as k-Nearest Neighbor, Local Outlier Factor, and Support Vector Data Description, for data mining in the context of classification. The authors provide insights into the strengths and weaknesses of each method through their evaluation using various datasets and performance metrics. Chawla et al. [5] conducted a comparison of several Local Outlier Factor (LOF) algorithms for outlier detection in data streams. The study analyzed the performance of LOF algorithms, including the original LOF, k-distance LOF, and Weighted LOF, in terms of accuracy, speed, and scalability using three real-world datasets. The results showed that the k-distance LOF algorithm is a suitable choice for outlier detection in data streams as it outperforms other LOF algorithms in terms of accuracy and speed. The study also highlights the importance of selecting an appropriate number of neighbors for the LOF algorithm, as it significantly impacts outlier detection performance.
Sakurada et al. [6] proposed a new approach to anomaly detection using autoencoders with nonlinear dimensionality reduction. The method trains an autoencoder network to learn normal patterns in the data and uses the learned model. methods in accuracy and robustness for anomaly detection.

Challenges of detecting outliers in high-dimensional data and compares conventional outlier detection methods such as Mahalanobis distance, k-expensive, while density-based methods such as Local Outlier Factor (LOF) and its variants are more efficient and effective in handling high-dimensional data. The paper highlights the importance of outlier detection in data mining, the challenges faced in detecting outliers in large and non-linear datasets, and how pattern-based techniques and weighted approaches can address these challenges.

Table 1

| Paper-Year | Author | Objective | Advantages | Limitations |
|---|---|---|---|---|
| MWFP-outlier: Maximal weighted frequent-pattern-based approach for detecting outliers from uncertain weighted data streams-2022 | Cai et al. [8] | Propose a new method for detecting outliers in uncertain weighted data streams by using maximal weighted frequent patterns | Effectively handle uncertain weight information in the data, improve the accuracy of outlier detection | Computational complexity, scalability for handling large datasets, Low accuracy over parameter-tuning. |
| Detecting Outliers from Pairwise Proximities: Proximity Isolation Forests-2023 | Mensi et al. [9] | Propose a new method for O.T in large datasets by using pairwise proximities and decision trees. | Able to handle high-dimensional data | Overfitting in high-dimensional datasets. |
| An efficient approach for outlier detection from uncertain data streams based on maximal frequent patterns-2020 | Cai et al. [10] | Propose a new method for detecting outliers in uncertain data streams by using maximal frequent patterns. | Use of frequent patterns can improve the accuracy of outlier detection compared to traditional methods. | May struggle to identify complex or non-linear relationships in the data. |
| A fuzzy proximity relation approach for outlier detection in the mixed dataset by using rough entropy-based weighted density method-2021 | Sangeetha et al. [11] | Propose a new method for detecting outliers in mixed datasets | Can effectively handle mixed datasets | Computational complexity of the algorithm and the potential for overfitting in high-dimensional datasets |
| An outlier mining-based method for anomaly detection -2007 | Wu et al. [12] | Propose a new method for detecting anomalies | Effectively handle large datasets | Can't perform well in non-linear relationships, struggle to identify anomalies in multivariate data |
| A novel weighted frequent pattern-based outlier detection method applied to data stream-2019 | Yuan et al. [13] | Propose A method for detecting outliers in data streams by using weighted frequent patterns | Improve the accuracy of outlier detection compared to traditional methods. | Overfitting in high-dimensional datasets |
| Multi-Hierarchy Attribute Relationship Mining Based Outlier Detection for Categorical Data-2019 | Hu et al. [14] | Propose A method to detect outliers in categorical data using multi-hierarchy attribute relationship mining techniques. | Effectively handle categorical data | Not perform well for datasets with complex structures or non-linear relationships |
| Outlier Detection using Clustering Techniques--K-means and K-median-2020 | Angelin et al. [15] | Proposed a method for detecting outliers by using clustering techniques, specifically K-means and K-median. | Improved accuracy of outlier detection compared to traditional method | May also struggle to identify outliers in datasets with non-uniformly distributed data |
| An efficient outlier detection approach on weighted data stream based on minimal rare pattern mining-2019 | Cai et al. [16] | presents a new method for outlier detection in weighted data streams | Low computational complexity and high Scalability | Incorrect choice of algorithm can be challenging in some cases., Difficulty in defining "rare patterns" |
| Weighted Outlier Detection of High-Dimensional Categorical Data Using Feature Grouping-2020 | Li et al. [17] | Presents a new method for outlier detection in high-dimensional categorical data. | Making it easier to understand the reason for a data point being identified as an outlier | Limited to categorical data |
| WMEVF: An outlier detection methods for categorical data-2016 | Rokhman et al. [18] | Eeffective approch in detecting outliers in categorical data | Suitable for datasets with many features. scalable, making it suitable for large datasets with many data points. | Limited to categorical data |

## V. FLOW DESIGN OF THE WEPAMADM-OUTLIER

As shown in Fig. 1 the Weighted Outlier Detection (WOD) model can be designed as a computer-scientific algorithm for identifying outliers in a dataset. The WOD model should be able to handle datasets with complex and heterogeneous distributions, as well as datasets with varying levels of relevance for different features.

### A. Problem Formation

Weighted Outlier Detection (WOD) is the problem of identifying data points that deviate significantly from the majority of the data in a dataset, considering the data points' importance (weights). The objective of WOD is to find instances that are significantly different from the rest of the data, where the significance is calculated based on the weight of each instance. Mathematically, WOD can be formulated as follows: Given a dataset X consisting of n instances $x_1, x_2, ..., x_n$ with associated weights $w_1, w_2, ..., w_n$, the goal is to find a $S_X$ such that the instances in S are significantly different from the majority of the instances in X, based on some distance measure and considering the instance weights. The WOD problem is relevant in various domains, including but not limited to finance, healthcare, and marketing, where it is important to identify important data points that may have a significant impact on the analysis of the data.

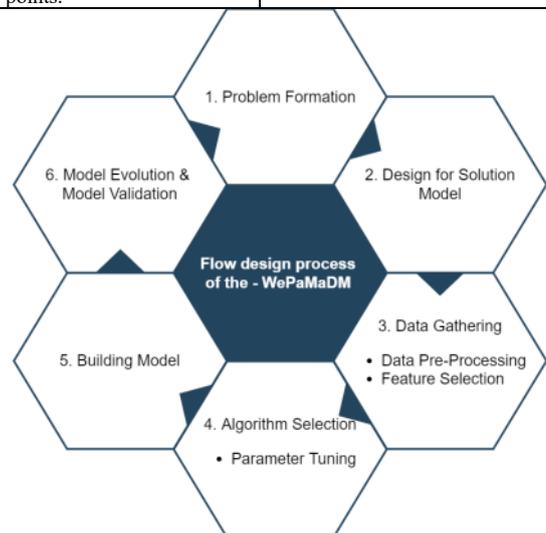

Fig 1 Flow design process of the - WePaMaDM

*B. Design of Solution Model*

The solution model for the Weighted Outlier Detection (WOD) system can be designed as follows:
1. Preprocessing & Distance measure: Clean the dataset and remove any missing or irrelevant information. Normalize the data if necessary. Choose a suitable distance measure to quantify the difference between instances. This could be Euclidean distance, Mahalanobis distance, or any other appropriate measure.
2. Assign weights to instances based on importance, using domain knowledge or statistical methods. Calculate outlier score based on distance measure and weights, reflecting deviation from the majority of data.
3. Threshold selection: Select a suitable threshold for the outlier score, which will determine which instances are considered outliers. This could be done using statistical methods or by setting a fixed value.
4. Outlier identification and Visualization: Identify the instances with outlier scores above the threshold as outliers.

*C. Data Gathering and Pre-processing*

Data Gathering and Data Preprocessing for the Weighted Outlier Detection (WOD) system can be designed as follows:
1. Data Gathering: Obtain the dataset from various sources, including but not limited to databases, file systems, and APIs. Ensure that the data is complete, consistent, and relevant to the problem at hand.
2. Data Cleaning Normalization: Clean the data to remove missing, irrelevant, or inconsistent information. This could include removing duplicates, imputing missing values, or transforming categorical variables.
3. Data Transformation: Transform the data if necessary to make it suitable for the WOD problem. This could include reducing the dimensionality of the data, transforming the data into a different representation, or converting the data into a more suitable format.
4. Data Splitting: Split the data into training and testing datasets, with a suitable ratio, to evaluate the performance of the WOD system.

*D. Algorithm selection and Parameter tuning*

Algorithm Selection and Parameter Tuning for the Weighted Outlier Detection (WOD) system can be designed as follows:
1. Algorithm Selection: Choose an appropriate algorithm for the WOD problem based on the nature of the data and the requirements of the problem. Common algorithms for WOD include Density-Based Outlier Detection, Distance-Based Outlier Detection, and Angle-Based Outlier Detection.
2. Parameter Setting: Set the parameters of the chosen algorithm based on the data characteristics and the desired performance. This could include setting the distance measure, the weight assignment method, the threshold for the outlier score, and the number of nearest neighbors used in the algorithm.
3. Hyperparameter Tuning: Tune the hyperparameters of the algorithm to optimize its performance. This could be done using grid search, random search, or Bayesian optimization. The objective of hyperparameter tuning is to find the optimal combination of hyperparameters that results in the best performance of the algorithm.
4. Model Selection: Select the best-performing algorithm and its optimal parameters based on the validation results.

*E. Building Model*

Building the Model for the Weighted Outlier Detection (WOD) system can be designed as follows:
1. Implementation: Implement the chosen algorithm with its optimal parameters in a suitable programming language, such as Python, R, or MATLAB. Ensure that the implementation is correct, efficient, and scalable.
2. Model Training: Train the WOD model on the training dataset obtained in the data preprocessing stage. The model should be trained to learn the patterns in the data and to identify outliers.
3. Model Testing: Test the WOD model on the testing dataset to evaluate its performance and robustness. The testing should be performed using the appropriate evaluation metrics, such as precision, recall, F1 score, and accuracy.
4. Model Validation and Deployment: Validate the WOD model using appropriate validation techniques, such as cross-validation, to ensure that the model is generalizable to new data.

*F. Model Evaluation and validation*

Evaluating the Model for the Weighted Outlier Detection (WOD) system can be designed as follows:

1. Model Testing: Test the WOD model on the testing dataset to evaluate its performance using the selected metrics. The testing should be performed on the out-of-sample data to ensure that the model generalizes well to new data.
2. Model Validation: Validate the WOD model using appropriate validation techniques, such as cross-validation, to ensure that the model is robust and generalizable. The validation should be performed on multiple partitions of the data to ensure that the model is consistent and reliable.
3. Model Comparison and Refinement: Compare the performance of the WOD model with other models or algorithms to ensure that it provides meaningful and accurate results. The comparison should be performed using the same evaluation metrics and data to ensure a fair comparison.

VI. PROPOSED ARCHITECTURE OF THE WEPAMADM-OUTLIER

As shown in Fig. 2, in the context of weighted outlier detection, the amount of data to be buffered is a crucial factor. This data buffer is used to store the input values that will be processed to determine the presence of outliers in the dataset. The size of the data buffer is directly proportional to the computational resources required to perform the outlier detection algorithms. Hence, it is important to strike a balance between the amount of data to be buffered and the computational resources available to ensure that the outlier detection process is performed efficiently. It is recommended to allocate an adequate amount of memory to the data buffer to ensure that the algorithm has sufficient data to perform a thorough analysis while avoiding memory overflow issues.

As a Second Stage, Data pre-processing is an important step to ensure accurate and reliable results. The purpose of data pre-processing is to clean and prepare the input data for analysis. This typically involves steps such as removing missing or corrupted data, normalizing the data to scale, and transforming the data into a suitable format for the outlier detection algorithm. Additionally, feature selection and dimensionality reduction techniques may be applied to remove irrelevant or redundant features from the dataset. The outcome of the data pre-processing step is a cleaned and transformed dataset that can be used for outlier detection with a high degree of accuracy. It is important to note that the success of the outlier detection process depends heavily on the quality of the data pre-processing step, as inaccurate or incomplete data can lead to false positive or false negative results.

The weight-based pattern clustering approach is a method used to group large amounts of data into meaningful clusters. This approach involves several steps including the initialization of centers and weights, weighted K-means clustering, center updates, and outlier detection. The first step of this approach is the initialization of centers and weights. Centers can be selected randomly from the data, or using methods such as k-means++. Weights are also assigned to each data point, reflecting the significance of each point.

Weight-based outlier detection, and the center and variance matrix of the data are important elements that contribute to the accuracy of the outlier detection process. The center of the data is typically represented by the mean of the data points and is used to calculate the distance between each data point and the center of the cluster. This distance is used to determine the density of each data point and, ultimately, whether it is an outlier. The variance matrix, on the other hand, is used to represent the spread of data around the center. This information is used to calculate the Mahalanobis distance, which takes into account the covariance structure of the data and is a more robust measure of distance compared to Euclidean distance. Both the center and variance matrix is updated in each iteration of the weight-based outlier detection process allowing for a a more accurate representation of the data and a better detection of outliers. The use of the center and variance matrix in combination with the weight-based outlier detection process allows for a more robust and accurate method of identifying outliers in large amounts of data.

Next, weighted K-means weighted clustering is performed using the initial centers. This method involves assigning each data point to the closest center based on its weight, and then updating the centers and weights based on the new cluster structure. This process is repeated until the cluster structure remains stable between iterations. After the data has been clustered, the centers are updated based on the weighted mean of the points assigned to each cluster. The weights of each data point are also re-evaluated based on the new cluster structure. Sometimes, outliers in the data can be detected using the minPts method. This method involves calculating the density of each data point and then identifying those points with a density lower than a predefined threshold as outliers. These outliers can then be excluded from further analysis.

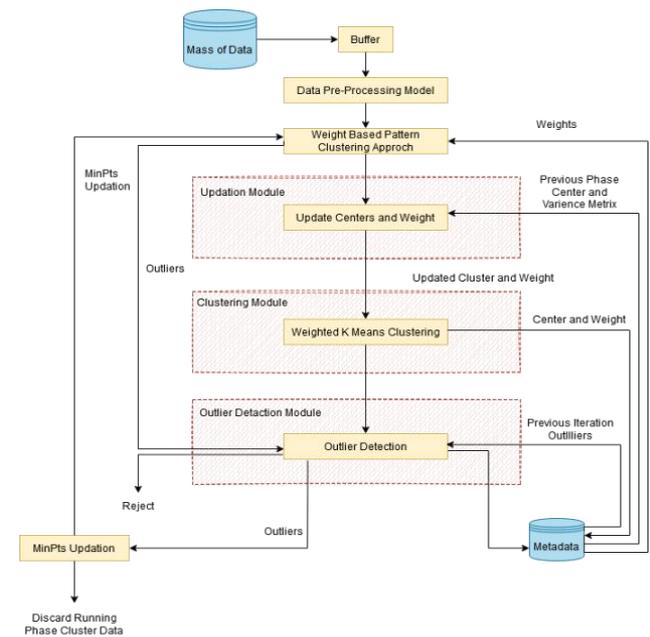

Fig. 2. WePaMaDM using pattern approaches

## VII. RESULTS AND DISCUSSION

The proposed approach is a valuable contribution to the field of outlier detection, as it provides a more effective and efficient solution for dealing with large-scale datasets. The approach has the potential to be applied in various domains, such as intrusion detection, fraud detection, and anomaly detection. The results of this study show that the WePaMaDM-Outlier Detection approach is an effective solution for outlier detection in large-scale datasets. Evaluate the effectiveness of each method by calculating standard metrics such as precision, recall, F1 score, area under the receiver operating characteristic (ROC) curve, and the detection rate. The proposed method offers a significant improvement in terms of accuracy and computational efficiency compared to conventional outlier detection methods and has the potential to be applied in various real-world applications.

## VIII. ADVANTAGES & DISADVANTAGES

By considering the relative importance of different and each data points, This can result in improved accuracy and meaningful outlier detections compared to traditional outlier detection methods that treat all data points equally. The implementation relatively straightforward as it can easily integrate with any domain-specific business logic, making it a highly flexible and versatile technique for outlier detection. This is an important advantage, as it allows organizations to tailor their outlier detection systems to fit their specific needs and requirements, and can lead to more effective and efficient solutions for identifying outliers in large and complex datasets.

Proposed methods is the increased computational complexity compared to traditional unweighted outlier detection methods. This is because the calculation of weights for each data point can be computationally expensive, especially when working with large datasets. Which can impact the overall complexity of the outlier detection process. Therefore, method may be less suitable for real-time applications where computational efficiency is a concern. Biased data can lead to uncertainty in the results of weighted outlier detection methods or Incorrect interpretation in computation of the weights can result in incorrect outlier detection, thus compromising the accuracy of the results.

## IX. CONCLUSION AND FUTURE WORK

In conclusion, the present study has demonstrated the utility of a pattern-based approach for weighted outlier detection in various datasets. By considering the unique characteristics of the data and the specific requirements of the outlier detection problem, this approach effectively identified outliers while reducing the likelihood of false positive detections. The results of this research indicate that the pattern-based approach can provide valuable insights into system faults, fraudulent activity, and interesting patterns within the data. Furthermore, the incorporation of weighting factors allows for a more precise detection of outliers, making this approach particularly useful in situations where the identification of outliers is of paramount importance. Overall, this study provides evidence for the effectiveness of the pattern-based approach to weighted outlier detection and its potential applicability in a wide range of domains.